\documentclass[sigconf,nonacm]{acmart}

\usepackage{xcolor}

\AtBeginDocument{%
  }

\setcopyright{cc}
\setcctype{by}
\copyrightyear{2026}
\acmDOI{}
\acmISBN{}
\makeatletter
\renewcommand{\@copyrightpermission}{%
  \copyright\ 2026 The authors.}
\makeatother

\usepackage{amsmath}
\usepackage{mathtools}
\usepackage{subcaption}
\usepackage[capitalize,noabbrev]{cleveref}

\usepackage{tikz}
\usetikzlibrary{arrows.meta,positioning,calc,fit,backgrounds,shapes.multipart,decorations.pathreplacing}

\AtEndPreamble{%
  \theoremstyle{acmplain}%
  \newtheorem{assumption}[theorem]{Assumption}%
  \theoremstyle{acmdefinition}%
  \newtheorem{remark}[theorem]{Remark}%
  \theoremstyle{acmplain}%
}

\begin{document}

\title{Probabilistic Limit Order Book Forecasting with Conditional Diffusion Models}

\author{Junyu Chen}
\authornote{These authors contributed equally to this research.}
\affiliation{%
  \department{Department of Economics, Econometrics \& Finance}
  \institution{University of Groningen}
  \city{Groningen}
  \country{The Netherlands}}

\author{Franklin Liu}
\authornotemark[1]
\affiliation{%
  \department{Department of Mathematics}
  \institution{University of Chicago}
  \city{Chicago}
  \state{Illinois}
  \country{USA}}

\author{Moshu Xu}
\authornotemark[1]
\correspondingauthor
\affiliation{%
  \department{Department of Statistics and Data Science}
  \institution{Tsinghua University}
  \city{Beijing}
  \country{China}}
\email{xms23@mails.tsinghua.edu.cn}

\author{Lijian Yang}
\affiliation{%
  \department{Department of Statistics and Data Science}
  \institution{Tsinghua University}
  \city{Beijing}
  \country{China}}

\renewcommand{\shortauthors}{Chen et al.}

\begin{abstract}
Modeling limit order book (LOB) dynamics is central to market forecasting and execution-strategy evaluation. While diffusion models achieve strong empirical performance, existing LOB snapshot models typically lack a learned recurrent encoder and formal statistical guarantees. We introduce \textbf{LOBGrad}, a conditional DDPM framework for probabilistic LOB forecasting that combines an autoregressive RNN encoder with a diffusion-based generator. In particular, we establish an excess-risk oracle inequality for the one-step DDPM noise-prediction objective under exponential mixing and regularity assumptions. We further show that, under contractive dynamics, the dependence on the conditioning-window length can be reduced to a dependence on the latent-state dimension. Experiments on BTC/USDT, GOOGL, and META show that LOBGrad improves on its encoder-free diffusion ablation across all reported queue metrics and obtains competitive results against other baselines. It also reproduces the key stylized facts of LOB microstructure, and yields reasonably calibrated queue-based predictive intervals.  A stylized downstream task illustrates the potential value of the generated forecasts.
\end{abstract}


\maketitle
\hypersetup{
  pdfauthor={Junyu Chen, Franklin Liu, Moshu Xu, Lijian Yang},
  pdfcreator={LaTeX},
  pdfsubject={Probabilistic limit order book forecasting},
  pdfkeywords={Limit order book, diffusion models, generative models, probabilistic forecasting, market microstructure}
}

\section{Introduction}

Limit order books (LOBs) are the main trading mechanism in modern electronic markets \citep{2013_Gould_et_al}. A Level-2 LOB records the aggregate volume of outstanding buy and sell limit orders at each visible price level. The queue profile provides information about local liquidity, the supply of immediacy, and the short-horizon price impact faced by traders \citep{2018_bouchaud_et_al}. Accordingly, modeling LOB dynamics is central to tasks such as market forecasting, execution-strategy evaluation, and counterfactual market analysis \citep{2025_Backhouse_et_al, 2025_Wang_and_Ventre}.

Denoising diffusion probabilistic models (DDPMs) \citep{2020_Ho_et_al, 2021_Song_et_al} have recently demonstrated strong empirical performance in time-series forecasting \citep{2021_Rasul_et_al, 2024_Chen_et_al} and can learn highly complex distributions \citep{2023_Chen_et_al, 2026_Yu_Yu}. These capabilities make DDPMs a natural choice for modeling limit order book dynamics \citep{2018_bouchaud_et_al, 2025_Backhouse_et_al, 2025_Wang_and_Ventre}. Despite their empirical success, existing diffusion-based models of LOB snapshots typically do not explicitly employ a recurrent encoder of past LOB snapshots. Moreover, for the broader class of DDPMs with recurrent encoders, a general theoretical understanding of how conditioning on recurrently encoded histories affects predictive performance remains unexplored.

In this work, we address both limitations by introducing \textbf{LOBGrad}, an encoder-conditioned DDPM for probabilistic LOB forecasting. Based on the \textit{TimeGrad} framework \citep{2021_Rasul_et_al}, LOBGrad integrates an autoregressive RNN encoder with a conditional diffusion-based generator. Specifically, the recurrent encoder maps past LOB snapshots to a lower-dimensional hidden state \citep{2007_Schafer_Zimmermann, 2019_Sirignano_Cont, 2020_Hanson_Raginsky}. This paper makes three contributions. (1) We adapt conditional diffusion forecasting to jointly predict queue depths and mid-price returns from recurrently encoded past LOB snapshots. (2) For the one-step DDPM noise-prediction objective, we establish an excess-risk rate under the stated mixing and regularity assumptions. Under a contractive hidden-state recurrence, the dimension appearing in the encoder-approximation rate is the hidden-state dimension $m$ rather than the flattened conditioning-window dimension $s d_y$. (3) We show that LOBGrad outperforms baseline models on LOB-Bench-inspired distributional metrics \citep{2025_Nagy_et_al} using Bitcoin data and U.S. equity data for Alphabet and Meta, and reproduces key microstructure stylized facts. We additionally demonstrate the practical utility of accurately forecasting LOB dynamics in a downstream execution task.

The remainder of this paper is organized as follows. Section~\ref{sec:related_work} reviews related work. Section~\ref{sec:methodology} introduces the LOB representation and our model architecture. Section~\ref{sec:theory} presents the theoretical results. Section~\ref{sec:experiment} reports the experimental evaluation. Section~\ref{sec:conclusion} concludes the paper.

\section{Related Work} \label{sec:related_work}

LOBGrad models the conditional distribution over future LOB states and can generate multiple plausible trajectories. This objective differs from classification-based or point-forecasting methods \citep{2024_Prata_et_al}. Our work extends the literature in the following two respects.

\paragraph{Generative models for LOB dynamics.} A growing body of work uses generative models to capture LOB dynamics. Early approaches employ GANs to generate order streams and learn conditional distributions of future LOB snapshots \citep{2021_Coletta_et_al, 2020_Li_et_al}. Notably, \cite{2023_Cont_et_al} propose a Wasserstein GAN that conditions on the current LOB snapshot and models the transition between consecutive LOB snapshots. Recent work has moved beyond GANs. \cite{2023_Nagy_et_al} uses a token-level autoregressive S5 model to generate message streams for a LOB simulator, while \cite{2025_Berti_et_al} introduces TRADES, a Transformer-based conditional diffusion model that generates individual orders conditioned on recent order flow. At the snapshot level, \cite{2025_Backhouse_et_al} represents LOB snapshots as images and apply image-based diffusion. DiffLOB \citep{2026_Wang_and_Ventre} models future LOB trajectories conditional on the observed history and market-regime variables describing trend, volatility, liquidity, and order-flow imbalance. Its conditioning setup is closely related to that of our no-encoder ablation. Despite their empirical success, none of these diffusion-based snapshot models employs an autoregressive recurrent encoder. Their conditioning relies on the current LOB snapshot, image-based spatial structure, or regime-level information rather than on a learned recurrent summary of the past LOB snapshots.

\paragraph{Theory for diffusion models.} Statistical theory for diffusion models has established approximation, estimation, and sampling guarantees, mainly for unconditional distribution learning \citep{2023_Chen_et_al_score,2023_Oko_et_al,2023_Benton_et_al, 2023_Lee_et_al,2024_Li_and_Yan}. Recent work also studies conditional diffusion models, including classifier-free guidance and conditional distribution estimation with observed covariates \citep{2024_Fu_et_al,2025_Tang_et_al}. These results typically assume independent samples and use feedforward networks conditioned on fixed covariates. Thus, they do not directly apply to our setting, where the conditioning variables are learned recurrent summaries of overlapping windows from a dependent time series. We address this problem by bounding the excess risk of the one-step DDPM $\varepsilon$-prediction objective under exponential mixing. 

\section{Methodology} \label{sec:methodology}

\begin{figure*}[t]
\centering
\includegraphics[width=0.9\textwidth]{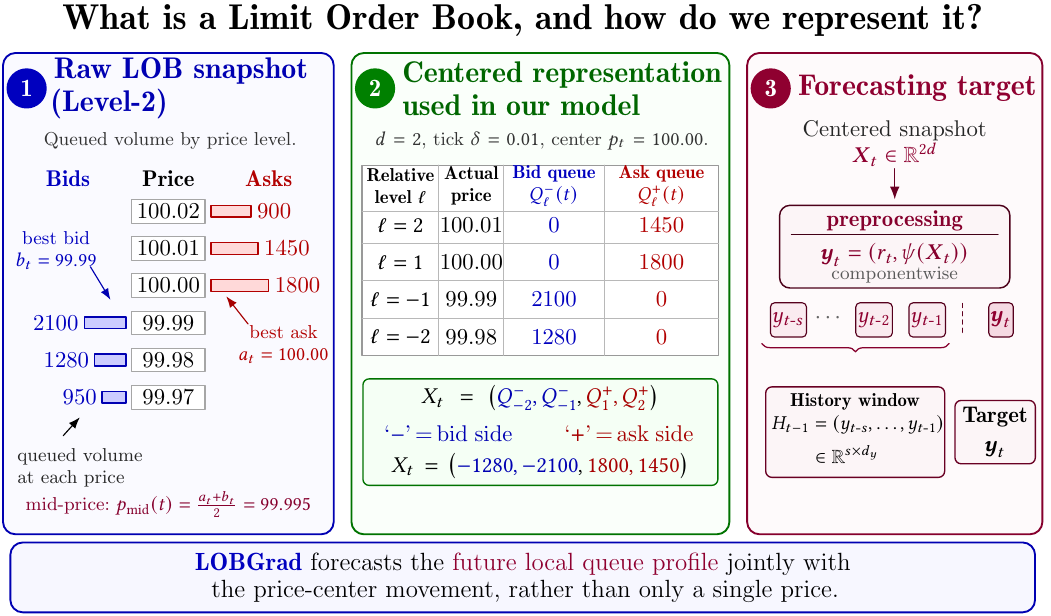}

\caption{Illustration of the raw Level-2 LOB, the centered representation used in our model, and the forecasting target.}
\Description{Bid and ask volumes are arranged around a price center, with negative values for bids and positive values for asks. The model uses past snapshots to predict the next price return and signed-log queue volumes.}
\label{fig:lob_explanation}
\end{figure*}

\subsection{Limit Order Book Representation}
\label{subsec:lob_rep}

We use Level-2 snapshots, where all orders at the same price are aggregated into a single queue size. Unlike Level-3 (message-level) data, Level-2 snapshots do not record individual order arrivals, cancellations, or modifications.

\paragraph{Centered order book.} Figure~\ref{fig:lob_explanation} illustrates our representation of Level-2 LOB snapshots. At time $t$, let $a_t$ and $b_t$ denote the best ask and best bid, respectively; that is, $a_t$ is the lowest price with positive sell volume, while $b_t$ is the highest price with positive buy volume. The corresponding mid-price is $p_{\mathrm{mid}}(t)=(a_t+b_t)/2$. We center the order book at the mid-price rounded to the nearest tick, and retain $d$ price levels on each side of the book. Let $p_t$ denote the actual price at time $t$ and $\delta$ denote the tick size. For $j=1,\ldots,d$, we define $Q^-_{-j}(t)$ as the bid volume at price $p_t-j\delta$, stored with a negative sign, and $Q^+_j(t)$ as the ask volume at price $p_t+(j-1)\delta$, stored with a positive sign. Empty queues are set to zero. The centered queue vector is
\begin{equation} \label{eq:centered_lob}
    X_t = \Bigl( \underbracket{ Q^{-}_{-d}(t),\ldots,Q^{-}_{-1}(t)}_{\text{bid queues}},\;\underbracket{Q^{+}_{1}(t),\ldots,Q^{+}_{d}(t)}_{\text{ask queues}} \Bigr) \in \mathbb{R}^{2d}.
\end{equation}
Let the return at time $t$ be $r_t=p_t-p_{t-1}$. The centered LOB representation is $z_t=(r_t,X_t)\in\mathbb{R}^{d_y}$, where $d_y=2d+1$ and $X_t$ is the $2d$-dimensional queue vector at time $t$.

\paragraph{Preprocessing.}
We apply the component-wise signed-log transformation $\psi(Q) = \operatorname{sign}(Q)\log(1+|Q|)$ to the queue vector, following \citep{2023_Cont_et_al} and extending their transformation to signed queue volumes. We then clip the transformed queue features and the return before training to prevent rare extreme values from destabilizing the optimization process. Throughout the main text and theory,
\begin{align*}
    y_t = \left( r_t, \psi(X_t) \right) \in \mathbb R^{d_y}
\end{align*}
denotes the transformed LOB representation, consisting of the price return and the transformed queue vector. Given a look-back length $s$, the observed history at time $t$ is
\begin{equation} \label{eq:history_window}
    H_{t-1} = (y_{t-s},\ldots,y_{t-1}) \in (\mathbb{R}^{d_y})^s .
\end{equation}
The forecasting target is the conditional distribution of $y_t$ given $H_{t-1}$.

\subsection{Encoder-conditioned DDPM} \label{subsec:ddpm}
LOBGrad follows the TimeGrad framework \citep{2021_Rasul_et_al} and consists of two learnable components: a time-series encoder $g^{(\mathrm{enc})}$ and a noise-prediction network $g^{(\epsilon)}$. It first compresses the observed past LOB snapshots into a recurrent latent state and then conditions each reverse-diffusion step on this state.

\paragraph{Autoregressive recurrent encoder.}
The encoder maps the look-back window $H_{t-1}$, defined in (\ref{eq:history_window}), to an $m$-dimensional conditioning state:
\begin{equation} \label{eq:encoder_map}
    h_{t-1} = g^{(\mathrm{enc})}(H_{t-1}) \quad \in [-1,1]^m.
\end{equation}
We let $g^{(\mathrm{enc})} \in \mathcal{G}^{(\mathrm{enc})}_{V,L,\kappa}(s,m,d_y)$, where $\mathcal{G}^{(\mathrm{enc})}_{V,L,\kappa}$ denotes the encoder function class. Here, $V$ and $L$ denote its width and depth, respectively, and $\kappa$ bounds its parameters. This class will be used to derive approximation and metric-entropy bounds in Section~\ref{sec:theory}. The encoder can be implemented using an RNN such as an LSTM.

\paragraph{Noise-prediction network.}
The denoising head $g^{(\epsilon)}$ is an MLP that takes as input the recurrent latent state $h_{t-1}$, the noised target $y_t^{(k)}$, and the diffusion step $k$:
\begin{equation}
\label{eq:epsnet_map}
    g^{(\epsilon)}: [-1,1]^m \times \mathbb{R}^{d_y} \times \{1,\ldots,K\} \to \mathbb{R}^{d_y}.
\end{equation}
We let $g^{(\epsilon)} \in \mathcal{G}^{(\epsilon)}_{D, W, \kappa}(m, d_y, K)$, where $\mathcal{G}^{(\epsilon)}_{D,W,\kappa}$ denotes the function class of the noise-prediction network. Here, $D$ denotes its depth, $W$ denotes its maximal hidden-layer width, and $\kappa$ bounds its affine weights and biases. This class will be used for the theoretical analysis in Section~\ref{sec:theory}.

\paragraph{Composite predictor.}
The complete predictor used in the DDPM is the composition of the autoregressive recurrent encoder and the noise-prediction network:
\begin{equation} \label{eq:tg_predictor}
    g^{(\mathrm{tg})}(H_{t-1},y_t^{(k)},k) := g^{(\epsilon)} \left( g^{(\mathrm{enc})}(H_{t-1}), y_t^{(k)}, k \right).
\end{equation}
Its corresponding composite function class $\mathcal{G}^{(\mathrm{tg})}_{V, L, D, W,\kappa}(s,m,d_y,K)$ is defined as 
\begin{equation} \label{eq:tg_class_method}
    \Bigl\{
        g^{(\mathrm{tg})} : g^{(\mathrm{enc})} \in \mathcal{G}^{(\mathrm{enc})}_{V,L,\kappa}(s, m, d_y),
        \, g^{(\epsilon)} \in \mathcal{G}^{(\epsilon)}_{D,W,\kappa}(m, d_y, K) \Bigr\}.
\end{equation}
For the theoretical analysis in Section~\ref{sec:theory}, we use a coordinate-wise clipped version of $g^{(\mathrm{tg})}$, with a clipping level proportional to $1+\|y_t^{(k)}\|_\infty$, to control the growth of the predictor and derive uniform empirical-process bounds.

\paragraph{Forward and reverse diffusion processes.}
Let $y_t^{(0)}=y_t$. Given a fixed variance schedule $\{\beta_k\}_{k=1}^K$ satisfying $0<\beta_k<1$, define $\alpha_k=1-\beta_k$ and $\bar{\alpha}_k=\prod_{j=1}^k\alpha_j$. The forward process generates the noised target at diffusion step $k$ as
\begin{equation} \label{eq:forward_reparam}
    y_t^{(k)} = \sqrt{\bar{\alpha}_k}\,y_t + \sqrt{1-\bar{\alpha}_k}\,\varepsilon_t, \qquad \varepsilon_t \sim \mathcal{N}(0,I_{d_y}).
\end{equation}
The reverse process is initialized with $y_t^{(K)}\sim\mathcal{N}(0,I_{d_y})$ and conditioned on the observed history through $g^{(\mathrm{tg})}$. For $k=K,\ldots,2$, its transition is
\begin{equation} \label{eq:reverse_transition}
    \begin{split}
        &p\bigl(y_t^{(k-1)}\mid y_t^{(k)},H_{t-1}\bigr) \\ 
        =\, & \mathcal{N} \left( \frac{1}{\sqrt{\alpha_k}} \left[ y_t^{(k)} - \frac{1-\alpha_k}{\sqrt{1-\bar{\alpha}_k}} g^{(\mathrm{tg})}(H_{t-1},y_t^{(k)},k) \right], \beta_k I_{d_y} \right).
    \end{split}
\end{equation}
At $k=1$, we use the reverse mean without adding Gaussian noise and set $\widehat y_t=y_t^{(0)}$. The composite predictor $g^{(\mathrm{tg})}$ is learned to estimate the injected Gaussian noise from the noised target, the diffusion step, and the LOB history. Its statistical properties are analyzed in Section~\ref{sec:theory}.

\subsection{Training and inference} \label{subsec:training}

\paragraph{Training objective.}
From a single transformed LOB series, we extract overlapping pairs $\{(H_{t-1},y_t)\}_{t=s+1}^T$. For simplicity, we still index them as $t=1, \dots, T$. Note that $T$ here is effectively the original sample size after removing the first $s$ observations. For each $t$, draw $N_t\sim\mathrm{Unif}\{1,\ldots,K\}$ and $\varepsilon_t\sim\mathcal N(0,I_{d_y})$, independently of the LOB process, and form $y_t^{(N_t)}$. For any network $g \in \mathcal{G}^{(\mathrm{tg})}_{V,L, D, W, \kappa}$, define empirical and population denoising risks as
\begin{align}
    \label{eq:empirical_risk_method}
    \mathcal{R}_T(g) &:= \frac{1}{T} \sum_{t=1}^T \left\| \varepsilon_t - g(H_{t-1},y_t^{(N_t)},N_t) \right\|_2^2, \\
    \mathcal{R}(g) &:= \mathbb{E} \left[ \left\| \varepsilon_t - g(H_{t-1},y_t^{(N_t)},N_t) \right\|_2^2 \right].
\end{align}
LOBGrad estimates the conditional denoiser by empirical risk minimization
\begin{equation}
\label{eq:erm_method}
    \widehat g_T^{(\mathrm{tg})}
    \in
    \operatorname{argmin}_{g\in
    \mathcal{G}^{(\mathrm{tg})}_{V,L,D,W,\kappa}(s,m,d_y,K)}
    \mathcal{R}_T(g).
\end{equation}
In practice, \eqref{eq:erm_method} is optimized by stochastic gradient descent (SGD), with one Monte Carlo draw of \((N_t,\varepsilon_t)\) for each training example at each optimization step. 

\paragraph{Inference.}
Given a new observed LOB history $H_{t-1}$, we generate a one-step forecast by initializing $y_t^{(K)}\sim\mathcal{N}(0,I_{d_y})$ and iterating the reverse update for \(k=K,\ldots,1\) according to
\begin{equation} \label{eq:inference_update_method}
    y_t^{(k-1)} = \tfrac{1}{\sqrt{\alpha_k}} \left[ y_t^{(k)} - \tfrac{1-\alpha_k}{\sqrt{1-\bar\alpha_k}}\, \widehat g_T^{(\mathrm{tg})}(H_{t-1},y_t^{(k)},k) \right] + \sqrt{\beta_k}\,z_k,
\end{equation}
where \(z_k\sim\mathcal N(0,I_{d_y})\) for \(k>1\) and \(z_1=0\). The terminal value $y_t^{(0)}$ is a draw from the learned conditional distribution of the next transformed LOB representation. Multi-step paths are obtained by appending generated states to the history window and repeating the same one-step sampler autoregressively. To recover a raw LOB snapshot, we apply the inverse preprocessing.

\section{Theoretical Guarantees} \label{sec:theory}
\paragraph{Notation.} Throughout this section, $T$ denotes the sample size, $\|\cdot\|_\infty$ the entrywise sup-norm, and $\|\cdot\|_2$ the Euclidean norm. We write $a_T \lesssim b_T$ if $a_T \le C\, b_T$ for a constant $C>0$, and $a_T \asymp b_T$ if $a_T \lesssim b_T \lesssim a_T$. The notations $\widetilde O(\cdot)$ and $\widetilde O_{\mathbb{P}}(\cdot)$ denote the standard $O(\cdot)$ and $O_{\mathbb{P}}(\cdot)$ orders up to factors polynomial in $\log T$. For a vector-valued function $f$ on a set $\mathcal A$, we write $\|f\|_{\infty,\mathcal A} := \sup_{x\in\mathcal A} \|f(x)\|_\infty$.

This section establishes the statistical guarantees for the DDPM $\varepsilon$-prediction risk. The main proof arguments are summarized in Remark~\ref{rem:proof_sketch}. Let $(H,Y)$ have the stationary distribution of $(H_{t-1},y_t)$. Independently of $(H,Y)$, let $N\sim\mathrm{Unif}\{1,\ldots,K\}$, $\varepsilon\sim\mathcal N(0,I_{d_y})$, and define $Y^{(N)} = \sqrt{\bar\alpha_N}\,Y + \sqrt{1-\bar\alpha_N} \varepsilon$. The oracle noise predictor is
\begin{equation} \label{eq:bayes_denoiser}
    g^\star(H,Y^{(k)},k) = \mathbb E \left[\varepsilon\mid H,Y^{(k)},k\right].
\end{equation}
For any measurable predictor $g$ with finite risk, the excess risk is \(\mathcal R(g) -\mathcal R(g^\star)\). By the projection identity for squared loss,
\begin{equation} \label{eq:excess_identity}
    \mathcal R(g) - \mathcal R(g^\star)
        =\; \mathbb E\left[ \left\|g(H,Y^{(N)},N)-g^\star(H,Y^{(N)},N)\right\|_2^2 \right].
\end{equation}
Thus, consistency for the DDPM objective is equivalent to estimating the conditional mean of the injected noise. Under the assumptions and sampling schemes of existing diffusion convergence results \citep{2023_Chen_et_al, 2023_Lee_et_al}, smaller score-estimation error yields tighter bounds on the discrepancy between generated and target distributions, with additional contributions from initialization and discretization errors.

\begin{assumption}[Statistical structure]
\label{ass:main_theory}
The transformed LOB process $(y_t)_{t\in\mathbb Z}$ is stationary and exponentially $\alpha$-mixing. There exists a constant \(B_y<\infty\) such that $\|y_t\|_\infty \le B_y$ almost surely. Let \(H=(y_{t-s},\ldots,y_{t-1})\) be the history. There exist an \(m\)-dimensional summary map $\Gamma^\dagger:([-B_y,B_y]^{d_y})^s \to [-1,1]^m$ and a reduced oracle map $\bar g^\star: [-1,1]^m\times \mathbb R^{d_y}\times\{1,\ldots,K\} \to \mathbb R^{d_y}$ such that the oracle noise predictor factors as
\begin{equation} \label{eq:latent_factorization}
    g^\star(H,Y^{(k)},k) = \bar g^\star\!\left(\Gamma^\dagger(H),Y^{(k)},k\right).
\end{equation}
The map $\Gamma^\dagger$ is $\gamma$-H\"older on $([-B_y,B_y]^{d_y})^s$. For each $k$, $\bar g^\star(\cdot,\cdot, k)$ is $\beta$-H\"older on every compact set $[-1,1]^m \times [-B_y-r,B_y+r]^{d_y}$, with H\"older radius at most polynomial in $r$.
\end{assumption}

\begin{remark}
    The map \(\Gamma^\dagger\) is an ideal population encoder. It is the low-dimensional state of the history that is sufficient for the one-step DDPM noise-prediction objective. The factorization \eqref{eq:latent_factorization} means that, for the purpose of predicting the injected noise from \((H,Y^{(k)},k)\), the full history \(H\) enters the oracle only through the summary \(h=\Gamma^\dagger(H)\). The map \(\bar g^\star\) is the corresponding oracle noise predictor function written in the reduced variables \((h,Y^{(k)},k)\). Note that the bar does not denote averaging. The normalization \(\Gamma^\dagger(H)\in[-1,1]^m\) is not substantive and can be replaced by any bounded range with only fixed changes in constants. 
    For any $a>0$, write $a=q+\rho$, where $q$ is a nonnegative integer and $\rho \in (0,1]$. An $a$-H\"older function has bounded derivatives up to order $q$, and its derivatives of order $q$ are $\rho$-H\"older continuous.
    Thus, \(\beta\) and \(\gamma\) measure the smoothness of the functions \(\bar g^\star\) and  \(\Gamma^\dagger\), respectively. 

\end{remark}

To handle the unbounded Gaussian perturbation, we restrict attention to the event $\|\varepsilon\|_\infty \leq r_T$, where $r_T\asymp\sqrt{\log T}$. The contribution of the complement is absorbed into the logarithmic factors. Define
\begin{align}
    \label{eq:head_approx}
    A^{\mathrm{head}}&(D,W) := \\
     & \inf_{g^{(\epsilon)} \in \mathcal G^{(\epsilon)}_{D,W,\kappa}} \max_{1\le k\le K} \left\|g^{(\epsilon)}(\cdot,\cdot,k)-\bar g^\star(\cdot,\cdot,k)\right\|_{\infty, \Xi_{r_T}}^2, \notag\\ 
    \label{eq:enc_approx}
    \Delta_{\mathrm{enc}}&(V,L) := \\ 
    & \inf_{g^{(\mathrm{enc})} \in \mathcal G^{(\mathrm{enc})}_{V,L,\kappa}} \sup_{H\in([-B_y,B_y]^{d_y})^s} \left\|g^{(\mathrm{enc})}(H)-\Gamma^\dagger(H)\right\|_\infty, \notag
\end{align}
where $\Xi_r= [-1,1]^m \times [-B_y-r,B_y+r]^{d_y}$. Let $\beta_0=\min\{\beta,1\}$ and define the following network-complexity quantity
\begin{equation} \label{eq:entropy_complexity_main}
    \begin{split}
        \mathfrak S(V,L,D,W) := &\bigl((2s-1)V+W+d_y+1\bigr)^2 \\
        &\times(sL+D+2)^2\\
        &\times \log \bigl((V+W)(L+D+2)\bigr).
    \end{split}
\end{equation}

\begin{theorem}[Excess risk for encoder-conditioned DDPMs]
\label{thm:excess_risk}
Suppose Assumption~\ref{ass:main_theory} holds and $\widehat g_T^{(\mathrm{tg})}$ is a measurable global minimizer of the empirical risk in \eqref{eq:erm_method}. Then
\begin{equation} \label{eq:main_decomposition}
    \begin{split} 
        &\mathcal R(\widehat g_T^{(\mathrm{tg})})-\mathcal R(g^\star) \\
    =& \widetilde O_{\mathbb P}\Bigg( \sqrt{\frac{\mathfrak S(V,L,D,W)}{T}} + \frac{\mathfrak S(V,L,D,W)}{T} \\
    & + A^{\mathrm{head}}(D, W) + \Delta_{\mathrm{enc}}(V,L)^{2\beta_0} \Bigg).
    \end{split}
\end{equation}
For fixed $s,m,K$ and recurrent depth $L$ and noise-prediction network depth $D=O(\log W)$,
$\mathfrak S(V,L,D,W)=\widetilde O(V^2+W^2)$. 
\end{theorem}

\begin{remark}
The first two terms in \eqref{eq:main_decomposition} are the stochastic cost of empirical risk minimization under dependence. The last two terms are approximation errors: $A^{\mathrm{head}}$ is the error of the denoising MLP when the ideal summary is supplied, while $\Delta_{\mathrm{enc}}^{2\beta_0}$ is the additional error from replacing $\Gamma^\dagger(H)$ by a learned recurrent summary.
\end{remark}

Standard ReLU approximation bounds give, up to logarithmic factors,
\begin{align}
\label{eq:approx_rates_generic}
    A^{\mathrm{head}}(D,W)
    &\lesssim W^{-2\beta/(d_y+m)}, \notag\\
    \Delta_{\mathrm{enc}}(V,L_0)^{2\beta_0}
    &\lesssim V^{-4\gamma\beta_0/(s d_y)},
\end{align}
for fixed encoder depth $L_0$ and head depth schedule $D=O(\log W)$.

\begin{corollary}[Generic window rate]
\label{cor:generic_rate}
Under Assumption~\ref{ass:main_theory}, choose
\begin{equation}
\label{eq:balanced_widths_generic}
    W
    \asymp
    T^{\frac{d_y+m}{2(d_y+m+2\beta)}},
    \qquad
    V
    \asymp
    T^{\frac{s d_y}{2(s d_y+4\gamma\beta_0)}} .
\end{equation}
Then
\begin{equation}
\label{eq:generic_rate}
    \mathcal R(\widehat g_T^{(\mathrm{tg})})-\mathcal R(g^\star)
    =
    \widetilde O_{\mathbb P}\!\left(
        T^{-\frac{\beta}{d_y+m+2\beta}}
        +
        T^{-\frac{2\gamma\beta_0}{s d_y+4\gamma\beta_0}}
    \right).
\end{equation}
\end{corollary}

Corollary~\ref{cor:generic_rate} treats $H$ as a flattened vector in $\mathbb R^{s d_y}$, so the encoder term deteriorates with the look-back dimension. A sharper bound is available when the sufficient summary is generated by a stable $m$-dimensional recurrence. Specifically, let $H=(u_1,\ldots,u_s)$ and suppose
\begin{align}
\label{eq:contractive_state}
    & h_0^\dagger = 0, \quad h_j^\dagger=\Phi^\dagger(P^\dagger u_j+h_{j-1}^\dagger),
    \quad j=1,\ldots,s, \notag\\
    &\Gamma^\dagger(H)=h_s^\dagger,
\end{align}
where $P^\dagger: \mathbb R^{d_y} \to \mathbb R^m$ is a bounded linear map, $\Phi^\dagger$ is locally $\gamma$-H\"older, and $\|\Phi^\dagger(P^\dagger u+h)-\Phi^\dagger(P^\dagger u+h')\|_\infty\le\lambda\|h-h'\|_\infty$ for some $\lambda<1$.

\begin{corollary}[Hidden-state refinement]
\label{cor:hidden_state_rate}
Under Assumption~\ref{ass:main_theory} and \eqref{eq:contractive_state}, suppose estimation is performed over the state-space recurrent subclass in which the map $P$ and the recurrent transition are learned jointly, and let $\widehat g_{T,\mathrm{st}}^{(\mathrm{tg})}$ be the resulting ERM. Let $\Delta_{\mathrm{st}}$ denote the analogue of $\Delta_{\mathrm{enc}}$ over this subclass. The encoder approximation term becomes
\begin{equation}
\label{eq:state_encoder_rate}
    \Delta_{\mathrm{st}}(V,L_1)^{2\beta_0}
    \lesssim V^{-4\gamma\beta_0/m},
\end{equation}
up to logarithmic factors. With
\[
    W\asymp T^{\frac{d_y+m}{2(d_y+m+2\beta)}},
    \qquad
    V\asymp T^{\frac{m}{2(m+4\gamma\beta_0)}},
\]
we obtain
\begin{equation}
\label{eq:hidden_state_rate}
    \mathcal R(\widehat g_{T,\mathrm{st}}^{(\mathrm{tg})})-\mathcal R(g^\star)
    =
    \widetilde O_{\mathbb P}\!\left(
        T^{-\frac{\beta}{d_y+m+2\beta}}
        +
        T^{-\frac{2\gamma\beta_0}{m+4\gamma\beta_0}}
    \right).
\end{equation}
\end{corollary}

The refinement explains the statistical role of the recurrent encoder: under the contractive-state assumptions, the dimension in the encoder approximation rate is the hidden-state dimension $m$ rather than the flattened dimension $s d_y$. This refinement concerns the one-step DDPM $\varepsilon$-prediction risk. Multi-step LOB paths are obtained by autoregressive rollout and are evaluated empirically in Section~\ref{sec:experiment}.

\begin{remark}[Proof sketch]
\label{rem:proof_sketch}
We first use the projection identity in~\eqref{eq:excess_identity} to express the excess risk as the estimation error of the oracle noise predictor. Although the training pairs contain overlapping history windows, each pair depends on only a fixed number $s$ of consecutive LOB observations. Thus, appending the independently sampled diffusion steps and Gaussian noises preserves exponential mixing, up to constants depending on $s$. Then, we apply Gaussian localization to control the diffusion noise. A Gaussian tail bound and a union bound show that the complement of this event has negligible probability. On the localized event, the predictor and the squared loss have controlled envelopes. We approximate the composite network class by a finite cover. Unrolling the recurrent encoder over the fixed look-back window reduces its entropy analysis to that of a feedforward network, whose covering bound can be combined with that of the denoising head. We then apply a concentration inequality for exponentially mixing sequences to this finite cover, followed by the standard empirical-risk-minimization inequality, which yields the two stochastic terms in Theorem~\ref{thm:excess_risk}. The localization radius and the Gaussian tail remainder contribute only polylogarithmic or asymptotically negligible factors.

For the approximation error, we approximate $\Gamma^\dagger$ with the encoder and $\bar g^\star$ with the denoising head. The factorization in Equation~\eqref{eq:latent_factorization} and the H\"older continuity of $\bar g^\star$ separate the resulting error into $A^{\mathrm{head}}(D,W)$ and $\Delta_{\mathrm{enc}}(V,L)^{2\beta_0}$, where $\beta_0=\min\{\beta,1\}$. This truncation reflects that when $\beta>1$ the required perturbation bound follows from local Lipschitz continuity. Applying the ReLU approximation bounds in Equation~\eqref{eq:approx_rates_generic} and balancing the approximation and stochastic terms yields Corollary~\ref{cor:generic_rate}. 

Finally, under the contractive realization in Equation~\eqref{eq:contractive_state}, errors across recurrent updates form a bounded geometric series. The encoder approximation therefore reduces to approximating an $m$-dimensional transition map rather than a function on the flattened $s d_y$-dimensional history. Substituting Equation~\eqref{eq:state_encoder_rate} and balancing the terms yields Corollary~\ref{cor:hidden_state_rate}.
\end{remark}

\section{Experimental Results} \label{sec:experiment}
\paragraph{Datasets.} We evaluate LOBGrad on three Level-2 LOB datasets: BTC/USDT from Binance and GOOGL and META from Databento. We focus on BTC/USDT, as Bitcoin is a highly liquid cryptocurrency, and its continuous trading provides a dense high-frequency sample without scheduled overnight closures. In this section, we provide the most detailed analysis for BTC/USDT and additional results for GOOGL and META to assess whether the results extend to U.S. equities. The BTC/USDT dataset consists of snapshots collected at 5-second intervals from the Binance public REST endpoint between October 1 and October 21, 2023. It contains 308{,}826 snapshots and is split chronologically into 15 training days, 2 validation days, and 4 test days. We aggregate the first 100 raw price levels into $d=5$ bins per side using a grid width of $\delta=2$ USDT, giving $d_y=2d+1=11$ in (\ref{eq:centered_lob}). This choice of $d$ and $\delta$ covers approximately 98\% of the observed mid-price changes between consecutive snapshots. The GOOGL and META datasets use Databento's MBP-10 schema and cover July 1--23, 2026. We retain the same model architecture and hyperparameters across all three datasets and refit the model separately on each dataset\footnote{The code is available at \url{https://github.com/junyuchen-econ/LOBGrad}.}.

\paragraph{Ablations and baselines.} We compare LOBGrad against three ablations (LG-Q, LG-R, and LG-NE), four external baselines (CGAN, PTM, TransLOB, and DeepLOB), and two classical statistical models (Hawkes and Poisson). LG-Q and LG-R use queue-only and return-only targets, respectively, while LG-NE removes the recurrent encoder and conditions on the flattened history of dimension $sd_y=1100$, isolating the effect of recurrent summarization. The external baselines are a conditional Wasserstein GAN (CGAN)~\citep{2023_Cont_et_al}; PTM, our re-implementation of the image-format snapshot diffusion model proposed by \citet{2025_Backhouse_et_al}, for which public code is unavailable; TransLOB~\citep{2020_Wallbridge}, a transformer-based model; and DeepLOB~\citep{2019_Zhang_et_al}. We adapt TransLOB and DeepLOB from their original classification setting~\citep{2024_Prata_et_al} to our probabilistic forecasting framework.

\paragraph{Experimental setup.} Hyperparameters for LOBGrad and the learnable baselines are selected on the validation split using the \textit{Optuna} Python package with median pruning. The selected model uses a look-back length of $s=100$, a one-layer LSTM with hidden size $m=151$, a $255$-unit noise MLP, and a $32$-dimensional diffusion-step embedding. We use $K=102$ diffusion steps with a linear schedule from $\beta_1=10^{-4}$ to $\beta_K=0.1$. The model is trained with the Huber noise-prediction loss using Adam. The empirical study covers three components: distributional generation quality evaluated on 500 held-out windows with 100-step contexts, 20-step forecast horizons, and 100 autoregressive rollouts per window; an analysis of return stylized facts; and a downstream limit-order execution task. The models are trained on a single NVIDIA RTX~5080 GPU. LOBGrad has 248{,}181 parameters and generates 100 paths at one-step and 20-step forecast horizons in 0.038 and 0.37 seconds, respectively, on a single Apple M4 CPU. Both runtimes are below the 5-second snapshot interval, allowing forecasts to be regenerated at every snapshot without dedicated hardware.

\subsection{Generation Quality}
Following LOB-Bench~\citep{2025_Nagy_et_al}, we report Wasserstein-1 distances for best-level total queue volume, order-book imbalance, and the marginal return distribution. We complement these with queue-profile MAE, queue and absolute-return ACF MAE, CRPS, and calibration coverage rate.

On BTC/USDT, Table~\ref{tab:headline} shows that LOBGrad achieves the best score on four
of the five queue metrics. The only exception is Queue ACF MAE, where the queue-only
ablation LG-Q is marginally better, suggesting a small trade-off between modeling
queue--return dependence and preserving short-lag queue autocorrelation. On return
metrics, LG-R performs best on return Wasserstein and return CRPS, as expected from its
dedicated return-only target. Nevertheless, LOBGrad improves over the encoder-free
diffusion baseline LG-NE on all queue metrics across all assets. Since
LG-NE keeps the diffusion objective fixed but removes the recurrent summary
\(g^{(\mathrm{enc})}(H_{t-1})\), the gap in queue profile, ACF, and CRPS supports the
empirical value of recurrent summarization. This observation is consistent with
Corollary~\ref{cor:hidden_state_rate}. However, note that this is not a formal test of
the asymptotic bound, since Table~\ref{tab:headline} reports rollout metrics rather than the population DDPM $\varepsilon$-risk.

Table~\ref{tab:headline} also reports results for GOOGL and META, using the same
architecture and hyperparameters as the BTC/USDT model. On both equities, LOBGrad
achieves the best score on three of the five queue metrics, which generalizes the
BTC/USDT result to a different market and tick regime without re-tuning.

\begin{figure}[t] 
\centering 
\includegraphics[width=0.45\textwidth]{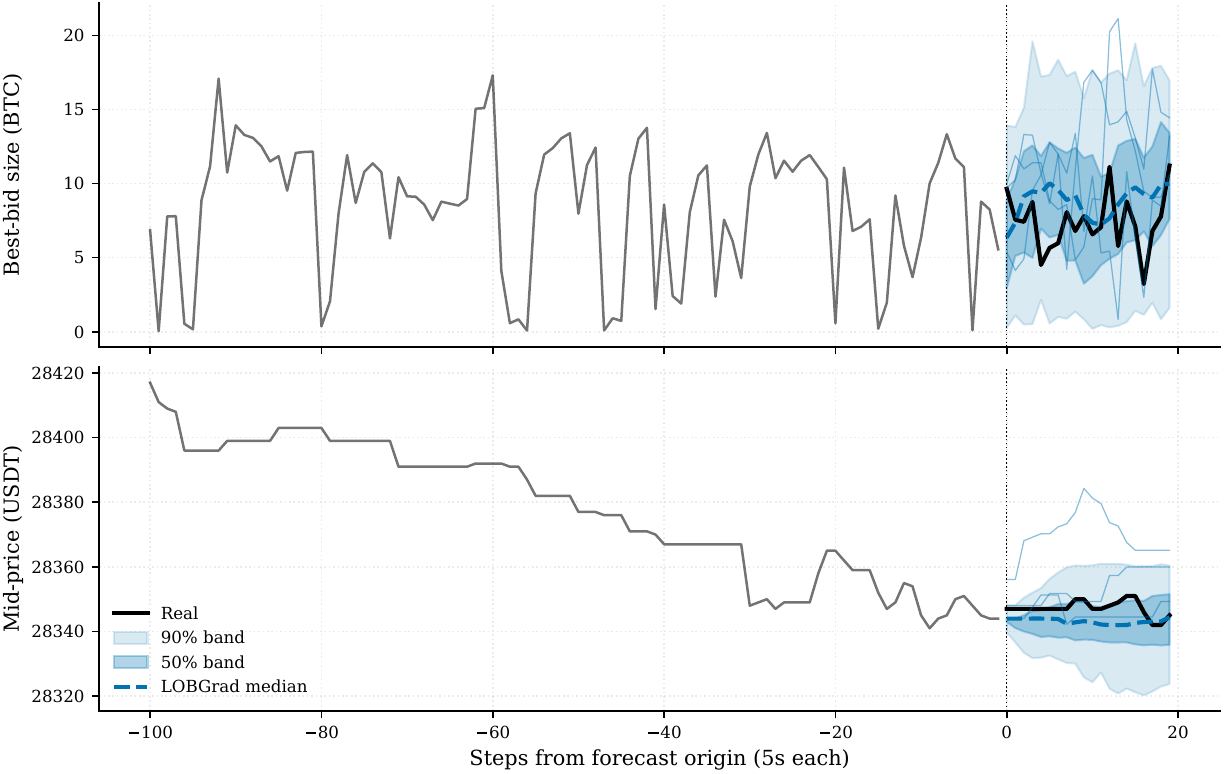} 
\caption{Sample best-bid queue (top) and mid-price (bottom) trajectories from LOBGrad with 50\% and 90\% predictive intervals on a held-out test window.} 
\Description{A dashed vertical line marks the forecast origin. The subsequent realized paths are compared with the forecast medians and shaded predictive intervals.}
\label{fig:trajectory} 
\end{figure}

Figure~\ref{fig:trajectory} illustrates the same conclusion at the path level. The predictive bands from LOBGrad cover the realized best-bid queue and mid-price trajectory in a held-out window. Table~\ref{tab:calibration_m500} reports direct interval-calibration diagnostics. We use $500$ rollouts per window to reduce Monte Carlo error. The 90\% intervals are close to nominal for queue-derived quantities, with empirical coverages \(0.876\) for all queue dimensions, \(0.892\) for best-level volume, and \(0.900\) for order-book imbalance. Return coverage is lower at \(0.844\), indicating that the model captures queue uncertainty more reliably than extreme return variation.

\begin{figure}[t] 
\centering 
\includegraphics[width=0.45\textwidth]{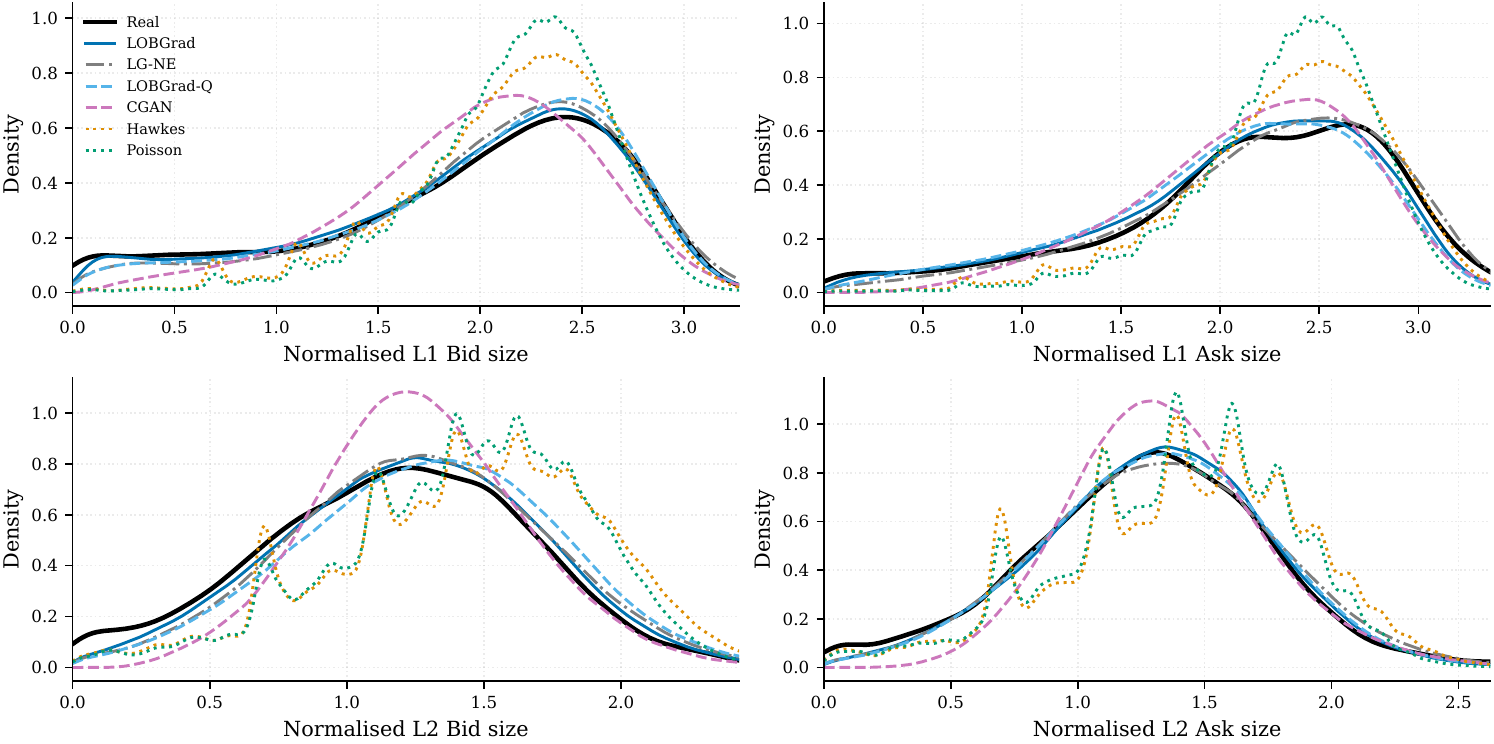} 
\caption{Marginal distributions of generated and real queue sizes at the top two book levels: bid (left) and ask (right), level 1 (top row) and level 2 (bottom row).} 
\Description{Four density plots compare LOBGrad and baseline models with held-out data. LOBGrad closely follows the main peaks and overall shapes of the empirical densities.}
\label{fig:marginal_distribution} 
\end{figure}

Figure~\ref{fig:marginal_distribution} shows the marginal distributions of queue size at the first two bid and ask levels, obtained by pooling samples over all 500 held-out windows and all 20 forecast steps. LOBGrad closely matches the empirical marginals.

\begin{table*}[t]
\caption{Distributional metrics over a $20$-step forecast horizon across 500 held-out windows. Lower values are better. \textbf{Bold} marks the best value per row. 
\textsc{ne} = no encoder; \textsc{q} = queue only; \textsc{r} = return only. Dashes indicate metrics that are not applicable.}
\label{tab:headline}
\centering
\footnotesize
\setlength{\tabcolsep}{4pt}
\begin{tabular*}{\linewidth}{@{\extracolsep{\fill}}l rrrr rr rr rr@{}}
\toprule
Metric & LOBGrad & LG-NE & LG-Q & LG-R & CGAN & PTM & TransLOB & DeepLOB & Hawkes & Poisson \\
\specialrule{0.9pt}{2pt}{2pt}
\multicolumn{11}{@{}l}{\bfseries BTC/USDT} \\
\cmidrule(l{0pt}r{0pt}){1-11}
\multicolumn{11}{@{}l}{\itshape Queue} \\
\quad L1 volume Wasserstein & \textbf{0.100} & 0.104 & 0.135 & --- & 0.197 & 0.126 & 0.201 & 0.176 & 0.266 & 0.330 \\
\quad OB imbalance Wasserstein & \textbf{0.095} & 0.107 & 0.096 & --- & 0.196 & 0.107 & 0.203 & 0.172 & 0.219 & 0.280 \\
\quad Queue profile MAE & \textbf{0.037} & 0.109 & 0.049 & --- & 0.044 & 0.140 & 0.068 & 0.082 & 0.098 & 0.100 \\
\quad Queue ACF MAE & 0.026 & 0.081 & \textbf{0.024} & --- & 0.034 & 0.039 & 0.075 & 0.072 & 0.225 & 0.219 \\
\quad CRPS (queue) & \textbf{0.261} & 0.276 & 0.262 & --- & 0.284 & 0.309 & 0.285 & 0.284 & 0.341 & 0.312 \\
\addlinespace[3pt]
\multicolumn{11}{@{}l}{\itshape Returns} \\
\quad Return Wasserstein & 0.097 & 0.211 & --- & \textbf{0.085} & 0.280 & 0.197 & 0.258 & 0.243 & --- & --- \\
\quad Absolute returns ACF MAE & 0.011 & \textbf{0.008} & --- & \textbf{0.008} & 0.017 & 0.012 & 0.013 & 0.014 & --- & --- \\
\quad CRPS (returns) & 1.719 & 1.760 & --- & \textbf{1.694} & 1.837 & 1.786 & 1.912 & 1.855 & --- & --- \\
\specialrule{0.9pt}{5pt}{2pt}
\multicolumn{11}{@{}l}{\bfseries GOOGL} \\
\cmidrule(l{0pt}r{0pt}){1-11}
\multicolumn{11}{@{}l}{\itshape Queue} \\
\quad L1 volume Wasserstein & 0.373 & 0.406 & \textbf{0.369} & --- & 0.461 & 0.528 & 0.519 & 0.418 & 0.492 & 0.553 \\
\quad OB imbalance Wasserstein & 0.230 & 0.367 & 0.292 & --- & 0.555 & 0.245 & \textbf{0.146} & 0.172 & 0.426 & 0.525 \\
\quad Queue profile MAE & \textbf{0.091} & 0.098 & 0.097 & --- & 0.101 & 0.111 & 0.146 & 0.108 & 0.340 & 0.349 \\
\quad Queue ACF MAE & \textbf{0.024} & 0.044 & 0.036 & --- & 0.353 & 0.035 & 0.045 & 0.046 & 0.362 & 0.349 \\
\quad CRPS (queue) & \textbf{0.536} & 0.547 & 0.543 & --- & 0.675 & 0.601 & 0.593 & 0.563 & 1.337 & 1.318 \\
\addlinespace[3pt]
\multicolumn{11}{@{}l}{\itshape Returns} \\
\quad Return Wasserstein & 0.049 & 0.061 & --- & 0.049 & 0.123 & 0.046 & 0.045 & \textbf{0.044} & --- & --- \\
\quad Absolute returns ACF MAE & 0.012 & 0.011 & --- & \textbf{0.006} & 0.168 & 0.016 & 0.013 & 0.013 & --- & --- \\
\quad CRPS (returns) & \textbf{4.307} & 4.397 & --- & 4.315 & 5.445 & 4.395 & 4.372 & 4.358 & --- & --- \\
\specialrule{0.9pt}{5pt}{2pt}
\multicolumn{11}{@{}l}{\bfseries META} \\
\cmidrule(l{0pt}r{0pt}){1-11}
\multicolumn{11}{@{}l}{\itshape Queue} \\
\quad L1 volume Wasserstein & \textbf{0.204} & 0.208 & 0.424 & --- & 0.457 & 0.338 & 0.370 & 0.376 & 0.640 & 0.658 \\
\quad OB imbalance Wasserstein & \textbf{0.128} & 0.232 & 0.261 & --- & 0.683 & 0.322 & 0.131 & 0.129 & 0.593 & 0.643 \\
\quad Queue profile MAE & 0.149 & 0.188 & 0.177 & --- & 0.154 & \textbf{0.145} & 0.154 & 0.157 & 0.332 & 0.341 \\
\quad Queue ACF MAE & 0.044 & 0.054 & \textbf{0.034} & --- & 0.238 & 0.044 & 0.063 & 0.053 & 0.296 & 0.257 \\
\quad CRPS (queue) & \textbf{0.778} & 0.784 & 0.810 & --- & 1.049 & 0.900 & 0.844 & 0.847 & 1.598 & 1.587 \\
\addlinespace[3pt]
\multicolumn{11}{@{}l}{\itshape Returns} \\
\quad Return Wasserstein & 0.092 & \textbf{0.054} & --- & 0.068 & 0.327 & 0.327 & 0.264 & 0.337 & --- & --- \\
\quad Absolute returns ACF MAE & 0.010 & 0.010 & --- & 0.012 & 0.133 & 0.017 & \textbf{0.009} & 0.010 & --- & --- \\
\quad CRPS (returns) & 3.499 & 3.535 & --- & \textbf{3.493} & 4.929 & 3.758 & 3.654 & 3.723 & --- & --- \\
\bottomrule
\end{tabular*}
\end{table*}

\begin{table}[t]
\caption{Calibration summary on BTC/USDT test windows using $M=500$ rollouts per window and a $20$-step forecast horizon. We report empirical coverages for nominal 50\% and 90\% central intervals, and the mean width of the 90\% interval.}
\label{tab:calibration_m500}
\centering
\small
\setlength{\tabcolsep}{5pt} 
\begin{tabular}{lccc}
\toprule
Target & 50\% coverage & 90\% coverage & Mean 90\% width \\
\midrule
Returns            & 0.630 & 0.844 & 8.684 \\
Queues ($2d=10$)       & 0.442 & 0.876 & 6.213 \\
Best-level volume  & 0.483 & 0.892 & 11.034 \\
OB imbalance       & 0.479 & 0.900 & 1.243 \\
\bottomrule
\end{tabular}
\end{table}

\subsection{Stylized Facts}
\begin{figure}[t]
    \centering
    \includegraphics[width=\linewidth]{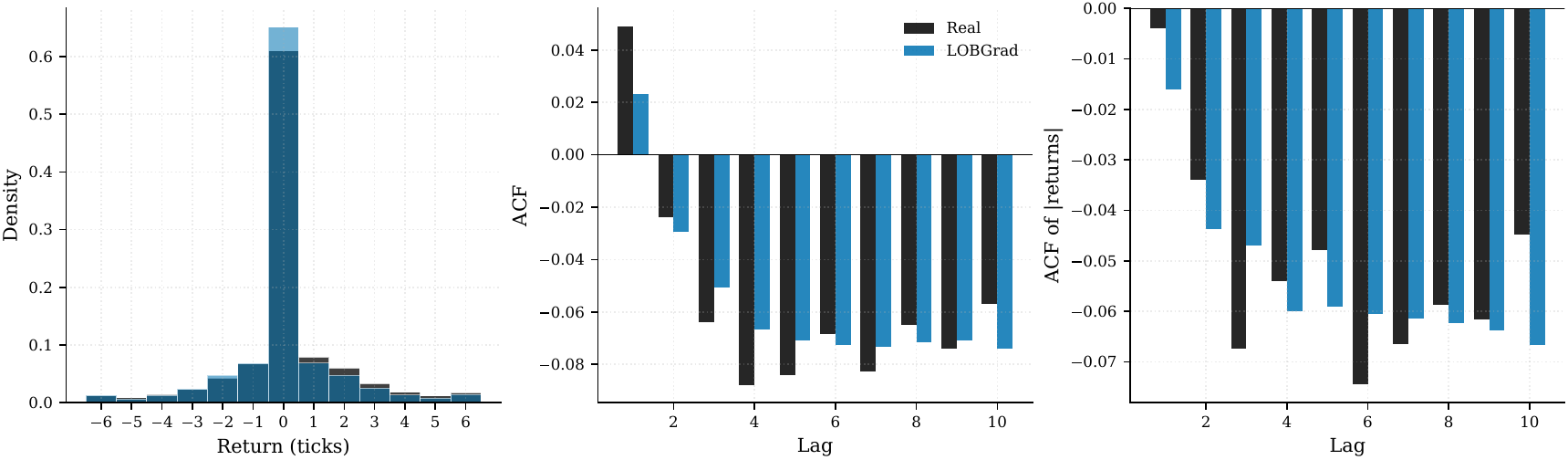}
    \caption{Stylized facts of 5-second BTC/USDT returns: real held-out data (black) versus LOBGrad samples (blue). Return distribution (left), the ACF of returns (middle), and the ACF of absolute returns (right).}
    \Description{Both return distributions peak sharply at zero. Return autocorrelation is positive at the first lag and negative at later displayed lags. Absolute-return autocorrelation is negative across the displayed lags.}
    \label{fig:price_stylized}
\end{figure}

Figure~\ref{fig:price_stylized} evaluates whether LOBGrad reproduces standard stylized facts of price series on BTC/USDT. We compare real and generated successive-snapshot mid-price changes using the marginal return distribution, the return ACF, and the absolute-return ACF. Generated samples are obtained from 100 autoregressive rollouts for each of the 500 evaluation windows. LOBGrad closely matches the empirical return distribution, captures the small first-lag autocorrelation in returns, and partially recovers the short-horizon dependence in absolute returns.

\subsection{Downstream Task}
Beyond distributional metrics, we test whether forecasts can be used for execution decisions. We consider a stylized single-shot sell problem. At the start of each test window, an agent must sell a quantity \(Q\) of BTC. The order size $Q$ is fixed across windows within a run and is twice the training-set mean cumulative bid depth over the first three grid bins. The action \(j=-1\) denotes an immediate market sell, while \(j\ge0\) posts a passive sell limit order at $p_{\mathrm{lim}} = p_{\mathrm{mid}}(0)+j\delta$, where $\delta=2$ USDT, $p_{\mathrm{mid}}(0)$ is the mid-price at order arrival, and waits up to 20 future snapshots. We take the offset as $j\in\{-1,0,1,2,3\}$. The order is treated as passively filled if the mid-price reaches the limit price or if the queue at the corresponding ask level is sufficiently depleted. Otherwise, it is executed at the final snapshot by walking the bid book. For each window, each model draws 100 autoregressive rollouts, simulates each candidate offset on each rollout, and selects the offset with the lowest mean simulated implementation shortfall $\mathrm{IS} = p_{\mathrm{mid}}(0)-\bar p_{\mathrm{fill}}$, where \(\bar p_{\mathrm{fill}}\) is the realized average fill price. We report mean IS and \(\mathrm{CVaR}_{95}(\mathrm{IS})\), the average shortfall over the worst 5\% of windows. The strategies for evaluation are: (i) static rules with fixed $j$; (ii) model-optimized selection and (iii) an oracle policy that uses the realized future path and is therefore an ex-post, non-implementable lower bound.

Table~\ref{tab:execution_limit_order} shows the mean--tail trade-off in this execution problem. The immediate market sell is always completed and therefore has the lowest non-oracle tail risk, but it pays the spread and has poor mean IS. Passive static offsets improve the mean when filled, but incur the cost of a terminal market order when unfilled. Among static rules, \(j=2\) attains the lowest mean IS at \(0.56\) USDT. The LOBGrad-optimized further reduces mean IS to \(0.03\) USDT and lowers \(\mathrm{CVaR}_{95}\) relative to the best-mean static rule, from \(47.86\) to \(41.69\) USDT. LOBGrad also outperforms CGAN, LG-NE, PTM, and DeepLOB on both metrics, while improving the mean IS compared to TransLOB at the cost of a slightly higher CVaR. This indicates that the generated rollouts contain useful information for selecting between aggressive and passive execution in individual windows.

\begin{table}[t]
\caption{Limit-order execution benchmark across 500 test windows, averaged over 5 seeds. IS per share in USDT (lower is better). Bold marks the best non-oracle value in each column.}
\label{tab:execution_limit_order}
\centering
\small
\renewcommand{\arraystretch}{1.25}
\setlength{\tabcolsep}{4pt}
\begin{tabular}{lrr}
\toprule
Strategy & Mean IS & $\mathrm{CVaR}_{95}(\mathrm{IS})$ \\
\midrule
$j={-1}$ (Market) & 5.89 & \textbf{8.45} \\
$j={0}$ & 1.75 & 38.90 \\
$j={1}$ & 0.75 & 43.42 \\
$j={2}$ & 0.56 & 47.86 \\
$j={3}$ & 2.93 & 51.51 \\
\midrule
CGAN-optimized & 0.19 & 47.35 \\
LG-NE-optimized & 0.28 & 43.24 \\
LOBGrad-optimized & \textbf{0.03} & 41.69 \\
PTM-optimized & 0.09 &  42.51 \\
TransLOB-optimized & 1.14 & \textbf{41.67} \\
DeepLOB-optimized & 0.59 & 41.92 \\
\midrule
Oracle & -6.73 & 7.07 \\
\bottomrule
\end{tabular}
\end{table}

\section{Conclusion} \label{sec:conclusion}

In this paper, we introduce LOBGrad, an encoder-conditioned DDPM that combines an autoregressive recurrent encoder with a conditional noise-prediction network for probabilistic LOB forecasting. Across BTC/USDT, GOOGL, and META, LOBGrad improves distributional forecasting performance, reproduces key microstructure stylized facts, and provides practical value in a downstream execution task. The theoretical analysis provides a one-step excess-risk bound for a recurrently conditioned DDPM under the stated mixing, smoothness, and network-class assumptions. This result motivates recurrent conditioning: under a stable contractive latent-state dynamics, the encoder approximation error term depends on the hidden-state dimension $m$ rather than the flattened history dimension $s d_y$. Future work will extend these guarantees to multi-step autoregressive forecasts and investigate recurrently conditioned diffusion models across broader market settings.

\bibliographystyle{ACM-Reference-Format}
\bibliography{ref}

\clearpage
\end{document}